\documentclass[10pt,twocolumn,letterpaper]{article}

\usepackage{cvpr}              

\usepackage{graphicx}%
\usepackage{multirow}%
\usepackage{amsmath,amssymb,amsfonts}%
\usepackage{amsthm}%
\usepackage{mathrsfs}%
\usepackage[title]{appendix}%
\usepackage{xcolor}%
\usepackage{textcomp}%
\usepackage{manyfoot}%
\usepackage{booktabs}%
\usepackage{algorithm}%
\usepackage{algorithmicx}%
\usepackage{algpseudocode}%
\usepackage{listings}%
\usepackage{physics}

\usepackage{xspace}
\makeatletter
\DeclareRobustCommand\onedot{\futurelet\@let@token\@onedot}
\def\@onedot{\ifx\@let@token.\else.\null\fi\xspace}

\def\eg{\emph{e.g}\onedot} 
\def\ie{\emph{i.e}\onedot}

\makeatother

\newcommand{\figref}[1]{Fig.~\ref{#1}}
\newcommand{\secref}[1]{Sec.~\ref{#1}}
\newcommand{\tabref}[1]{Table~\ref{#1}}

\definecolor{cvprblue}{rgb}{0.21,0.49,0.74}
\usepackage[pagebackref,breaklinks,colorlinks,citecolor=cvprblue]{hyperref}

\def\paperID{324} 
\def\confName{3DV\xspace}
\def\confYear{2024\xspace}

\title{RaNeuS: Ray-adaptive Neural Surface Reconstruction}

\author{
   Yida Wang\,\textsuperscript{1} \quad
   David Joseph Tan\,\textsuperscript{2} \quad
   Nassir Navab\,\textsuperscript{1} \quad
   Federico Tombari\,\textsuperscript{1,2} \\
   \textsuperscript{1}\,Technische Universität M\"unchen \enskip
   \textsuperscript{2}\,Google \\
}

\begin{document}
\makeatletter
\g@addto@macro\@maketitle{
\begin{figure}[H]
    \setlength{\linewidth}{\textwidth}
    \setlength{\hsize}{\textwidth}
    \centering
    \includegraphics[width=\textwidth]
    {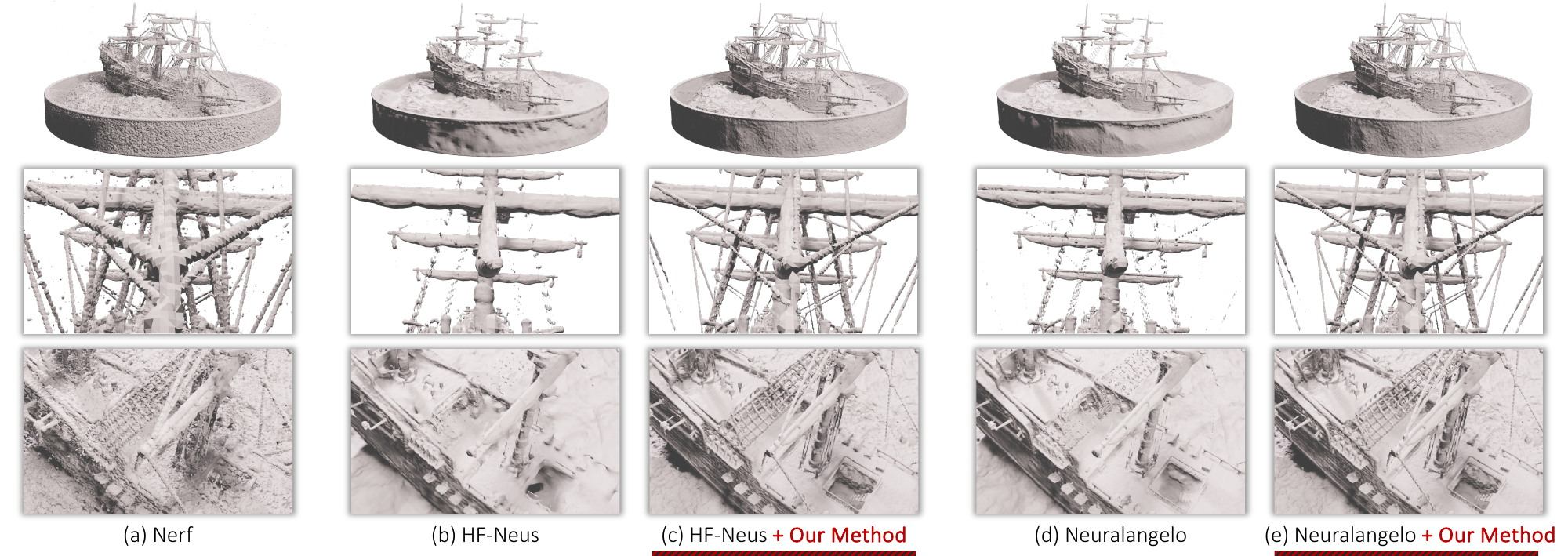}
    \caption{We improve the accuracy of the extracted surfaces $\mathbf{S}$ from signed distance field (SDF) based on follow-up works of NeuS~\cite{wang2021neus}, \eg HF-NeuS~\cite{wang2022hf} in (b) and Neuralangelo~\cite{li2023neuralangelo} in (d). Surfaces of all NeuS variants are extracted from zero-crossing space in SDF $f(\cdot)$, while surfaces of NeRF~\cite{mildenhall2021nerf} in (a) are extracted from a learned density field $\sigma(\cdot)$ with a threshold of 15. Our proposed method in (c) and (e) uncovers the lost details in NeuS-based methods with respect to (a), while keeping the smoothness in SDF compared to surfaces extracted from NeRF.}
    \label{fig:teaser}
\end{figure}
}
\maketitle

\begin{abstract}
Our objective is to leverage a differentiable radiance field \eg NeRF to reconstruct detailed 3D surfaces in addition to producing the standard novel view renderings.
There have been related methods that perform such tasks, usually by utilizing a signed distance field (SDF). However, the state-of-the-art approaches still fail to correctly reconstruct the small-scale details, such as the leaves, ropes,  and textile surfaces.
Considering that different methods formulate and optimize the projection from SDF to radiance field with a globally constant Eikonal regularization, we improve with a ray-wise weighting factor to prioritize the rendering and zero-crossing surface fitting on top of establishing a perfect SDF.
We propose to adaptively adjust the regularization on the signed distance field so that unsatisfying rendering rays won't enforce strong Eikonal regularization which is ineffective, and allow the gradients from regions with well-learned radiance to effectively back-propagated to the SDF.
Consequently, balancing the two objectives in order to generate accurate and detailed surfaces.
Additionally, concerning whether there is a geometric bias between the zero-crossing surface in SDF and rendering points in the radiance field, the projection becomes adjustable as well depending on different 3D locations during optimization.
Our proposed \textit{RaNeuS}
\footnote{Codes are released at \href{https://github.com/wangyida/ra-neus}{https://github.com/wangyida/ra-neus}.} are extensively evaluated on both synthetic and real datasets, achieving state-of-the-art results on both novel view synthesis and geometric reconstruction.
\end{abstract}

\section{Introduction}
\label{sec:intro}

Understanding the 3D structure from multi-view stereo (MVS) data performs well on multiple tasks, \eg dense reconstruction~\cite{furukawa2009accurate, kazhdan2013screened, gopi2000surface, cazals2006delaunay}, novel-view synthesis~\cite{mildenhall2021nerf, pumarola2021d, muller2022instant} and watertight surface reconstruction~\cite{wang2021neus, wang2022hf, wang2022neus2, zhao2022human}. 
Limiting the scope to reconstruction, MVS leverages the images collected from different camera positions to build a digital replica of a scene or an object.
Although traditional MVS reconstruction pipeline, \eg Poisson~\cite{kazhdan2013screened} or Delaunay~\cite{gopi2000surface, cazals2006delaunay}, performs well on both indoor and outdoor scenarios, they still have difficulty in reconstructing the finer details, \ie the surface is too noisy or even incomplete. 

More recently, there is a new trend in utilizing neural renderers~\cite{mildenhall2021nerf, pumarola2021d, muller2022instant} for reconstruction based on the Neural Radiance Fields (NeRF)~\cite{mildenhall2021nerf}. 
Leveraging the vast information learned in the radiance field, these works~\cite{wang2021neus, wang2022hf, wang2022neus2, zhao2022human, hou2023neudf, long2022neuraludf, guillard2022meshudf} aim at building the connections between the geometric implicit field, \eg signed distance field (SDF) or unsigned distance field (UDF), to the radiance field.
In this way, we can optimize them together so that we can easily extract the mesh directly from learned signed implicit field by matching cube~\cite{lorensen1987marching} to get watertight meshes. 

%


Determining how to merge the optimizations for novel view synthesis with the optimization for reconstruction is the real challenge. 
For example, we noticed a race condition in some cases wherein the SDF was optimized faster than the radiance field; consequently, limiting the influence of the radiance field in reconstruction.
Therefore, in this work, we focus on capturing the balance between optimizing the appearance and the geometry as shown in \figref{fig:teaser} in order to reconstruct fine-grained surfaces of a static scene or an object with the help of neural rendering. 


\section{Related Works}
\label{sec:related}

NeRF~\cite{mildenhall2021nerf} and its subsequent works ~\cite{pumarola2021d, barron2021mip, zhang2020nerf++} represent a remarkable advancement in novel view synthesis from multi-view stereo images. 
One disadvantage of training the radiance field is the low training and inference speed. Works introducing coding the input query point in a multi-resolution hash code~\cite{muller2022instant, zhao2022human, wang2022neus2} solves this problem, boosting the research on this topic.

The high-quality novel view rendering indicates a strong prospect for a detailed 3D geometric reconstruction, which is investigated by follow-up works~\cite{wang2021neus, long2022sparseneus, wang2022hf}. 
Common among these methods, the overall framework can be summarized with the three-stage pipeline: (1) camera pose estimation; (2) mesh initialization in signed implicit field; and, (3) efficient mesh refinement. 
Since the camera poses are usually given, typically computed through COLMAP~\cite{schoenberger2016sfm, schoenberger2016mvs} or HLOC~\cite{sarlin2019coarse}, the scope of our paper is associated with a cascaded pipeline with the step (2) to produce a high-quality mesh compared to the related works and then with the step (3) to refine the extracted mesh optimizing for efficiency and the textures on the mesh. 
In this paper, we observed that the accuracy of the proposed method surpasses the related work even without relying on mesh refinement.
The subsequent sections then focus on step (2) involving implicit mesh extraction with learned radiance field in \secref{sec:related_radiance} and effective training with hash encoding in \secref{sec:related_hashcodes}.

\subsection{Neural implicit surface}
\label{sec:related_radiance}

Extracting the mesh from a signed implicit field $f(\cdot)$, \textit{e.g.} the signed distance field (SDF)~\cite{oleynikova2016signed, zhang2021learning} or truncated SDF (TSDF)~\cite{izadi2011kinectfusion}, is determined by the values of the sampled positions at a zero threshold. Mathematically, the surface $\mathbf{S}$ of the scene can be obtained by extracting the zero-level set of the SDF.
%
Neural rendering~\cite{mildenhall2021nerf} can produce high-quality novel view synthesis, which we can exploit with the help of the learned radiance field. Extracting meshes from the radiance field such as NeuS~\cite{wang2021neus}, NeuS2~\cite{wang2022neus2}, Instant-NSR~\cite{zhao2022human}, and HF-NeuS~\cite{wang2022hf} has shown advantages against constructing meshes using traditional MVS fusion such as MeshMVS~\cite{shrestha2021meshmvs}.
As one of the first few works that extract mesh with the help of radiance field, VolSDF~\cite{yariv2021volume} applied the cumulative distribution function of Laplacian distribution to evaluate the density function from SDF for volume rendering and surface reconstruction. 
NeuralWarp~\cite{darmon2022improving} further improved the accuracy on low-textured areas by optimizing consistency between warped views of different images. 
Another series of works which are variants from 
NeuS~\cite{wang2021neus}, adopted an unbiased density function for signed distance field (SDF) produces more accurate surface reconstruction.
%
One of them is SparseNeuS~\cite{long2022sparseneus} which extends NeuS to use fewer images for reconstruction. 
Moreover,
HF-NeuS~\cite{wang2022hf} improves NeuS by proposing a simplified unbiased density function, and using hierarchical MLPs for detail reconstruction. 
Geo-NeuS~\cite{fu2022geo} incorporates structure-from-motion to add more constraints.
All of these methods learn SDFs, which can only reconstruct watertight models. 
In contrast, 
NeAT~\cite{meng2023neat} represents the 3D surface as a level set of SDF with a validity branch for estimating the surface existence probability at the query positions, which is proven to be helpful to reconstruct mesh for arbitrary topologies such as clothes in D3D~\cite{zhu2020deep} dataset.

Presenting the surface in an unsigned distance field~\cite{chibane2020neural} (UDF) is another solution to reconstruct arbitrary topologies. 
MeshUDF~\cite{guillard2022meshudf} is one of the first works proposing to extract mesh from a UDF, which can perform better in reconstructing non-watertight surfaces. 
More recently, learning UDF with the help of the radiance field is also practical, as long as the parameterized UDF is differentiable. NeuralUDF~\cite{long2022neuraludf} is used to learn in reconstructing 3D surfaces with arbitrary topologies. It adapts the density function of NeuS to UDFs by introducing an indicator function. However, this method can only learn highly-textured models due to the complex density function used in training. In contrast, NeUDF~\cite{hou2023neudf} proposes a UDF training method capable of reconstructing highly textured and textureless models without requiring masks. However, the problem of NeUDF is the biased rendering density field conditioned on the object surface. 

\subsection{Hash encoding}
\label{sec:related_hashcodes}
To solve the training efficiency issue in neural rendering, Instant-NGP~\cite{muller2022instant} proposes to use a hash encoding. 
Such an idea of encoding the input 3D positions $x \in \mathbb{R}^3$ into hash code using MLP for neural rendering generates an obvious advantage in converging speed compared to NeRF~\cite{mildenhall2021nerf} and NeuS~\cite{wang2021neus}.
But the meshes extracted from Instant-NGP do not reach the quality of neural rendering methods embedded with an implicit field as the intermediate feature space~\cite{wang2021neus, wang2022hf, zhao2022human} in terms of surface mesh reconstruction smoothness and precision. This is mainly because of the unknown threshold in extracting the surface mesh with a marching cube.
More recently, NeuS~2~\cite{wang2022neus2}, Instant-NSR~\cite{zhao2022human}, and Neuralangelo~\cite{li2023neuralangelo} were proposed to utilize multi-resolution 3D hash grids and neural surface rendering to achieve superior results in recovering dense 3D surface structures from multi-view images, enabling highly detailed large-scale scene reconstruction from RGB video captures.

\section{Methodology}
\label{sec:method}


Given a set of calibrated multi-view images capturing an object or a static scene with their corresponding camera poses, we jointly learn a structural surface $\mathbf{S}$ and the appearance $\mathbf{C}$ of the targeted scene through the appearance supervision~\cite{wang2021neus, long2022neuraludf, wang2022hf}. The learned set $\{ \mathbf{S}, \mathbf{C} \}$ are represented from the signed distance field (SDF) $f(x) : \mathbb{R}^3 \rightarrow \mathbb{R}$ where the value of each element is determined by a 3D position $x$, and a radiance field $c(x, v) : \mathbb{R}^3 \times \mathbb{S}^2 \rightarrow \mathbb{R}^3$ which is determined by both the position $x$ and the viewing direction $v \in \mathbb{S}^2$. Aiming at learning more precise zero-crossing surfaces in SDF by jointly training the SDF and the radiance field, we introduce two proposed factors $\lambda_{r}$ and $\lambda_{r}$ to make the SDF regularization more adaptive to improve the rendering quality and reduce the geometric bias.

\subsection{Neural rendering}

By enforcing the radiance supervision through the 2D image,
NeRF~\cite{mildenhall2021nerf} leverages volume rendering to match the ground truth for every camera pose with the rendered image. 
Specifically, we can generate the RGB for every pixel of an image by sampling $n$ points $\{~r(t_i) = o~+~t_i \cdot v~|~i = 1, \dots, n~\}$ along its camera ray $r$, where $o$ is the center of the camera, $t_i$ is the sampling interval along the ray and $v$ is the view direction. 
By accumulating the radiance field density $\sigma(r(t))$ and colors $c(r(t), v)$ of the sample points, we can compute the color $\mathbf{\hat{C}}$ of the ray as
\begin{align}
\mathbf{\hat{C}}(r) = \int_{t_\text{n}}^{t_\text{f}} T(t) \cdot \sigma(r(t)) \cdot c(r(t), v) ~ dt 
\label{equ:radiance}
\end{align}
where the transparency $T(t)$ is derived from the volume density $\sigma(r(t))$. $T(t)$ denotes the accumulated transmittance along the ray $r$ from the closest point $t_\text{n}$ to the farthest point $t_\text{f}$ such that
\begin{align}
T(t) = \exp \left( -\int_{t_\text{n}}^{t} \sigma(r(s)) ds \right)~.
\end{align}
Note that $T(t)$ is a monotonic decreasing function with a starting value $T(t_\text{n})$ of 1. 
The product $T(t) \cdot \sigma(r(t))$ is used as a weight $\omega(t)$ in the volume rendering of radiance in \eqref{equ:radiance}.

Since the rendering process is differentiable, our model can then
learn the radiance field $c$ from the multi-view images with the loss function that minimizes the color difference between the rendered pixels $\mathbf{\hat{C}}(r)$ with $i \in \{1, \dots, m\}$ and the corresponding ground truth pixels $\mathbf{C}(r)$ without 3D supervision as
\begin{align}
    \mathcal{L}_\text{rgb} = \frac{1}{m} \sum_{i=1}^{m} \left(
    \| \mathbf{\hat{C}}(r) - \mathbf{C}(r) \|_2 + 
    | \mathbf{\hat{C}}(r) - \mathbf{C}(r) |
    \right)
    \label{equ:loss_color}
\end{align} 
%
where $m$ denotes the batch size during training. 
Based on the same input and output, we would further investigate a way to implicitly learn a signed distance field $f$ to extract meshes embedded in \eqref{equ:radiance} during training.


\begin{figure*}[!ht]
\centering
\includegraphics[width=\linewidth]{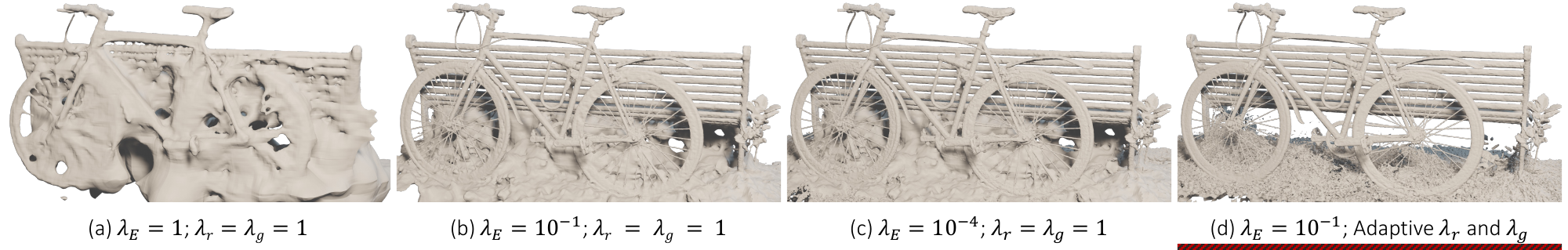}
\caption{Comparison of the Eikonal regularizer with adaptive factor $\lambda_\text{r}$ and $\lambda_\text{g}$ against constant Eikonal regularizer's weights on unbounded scene reconstruction in Mip-NeRF 360 dataset~\cite{barron2022mip}.}
\label{fig:ablation_weight}
\end{figure*}

\subsection{Ray-adaptive SDF optimization}



Aiming at extracting a 3D mesh from a region of interest in the neural rendering,
it is plausible to get a projection from a signed distance function (SDF) to the radiance field.
Here, we look for a function $\Phi$ that transforms the signed distance function so that it can be used to compute the density-related term $T(t) \sigma(r(t))$ in \eqref{equ:radiance}. 

We build our solution on top of HF-NeuS~\cite{wang2022hf}, where they set $\Phi(r(t))$ as the transparency $T(t)$.
Notably, the derivative of the transparency function $T(t)$ is the negative weighting function as
\begin{align}
\frac{d(T(t))}{dt} = -T(t) \sigma(r(t))~.
\label{equ:trans_density}
\end{align}
Given such formulation, 
the SDF surface lies on the maximum radiance weight.
The maxima is computed by setting the derivative of the weighting function to zero; thus, 
\begin{align}
\frac{d(T(t) \sigma(r(t)))}{dt} = -\frac{d^2(T(t))}{dt^2} = -\frac{d(T'(t))}{dt} = 0~.
\label{equ:unbias_weight}
\end{align}
%
%
To fulfill the criteria from \eqref{equ:trans_density} and \eqref{equ:unbias_weight}, HF-NeuS~\cite{wang2022hf} defined the transparency function $T_\text{s}(t)$ as the normalized sigmoid function $\left[1+\exp\left({s \cdot f(r(t))}\right)\right]^{-1}$
with a trainable parameter $s$. The scalar $s$ reveals how strong the SDF is related to the radiance field, which is usually increasing during training. In the initial stage of training, a small $s$ relaxes the connection between SDF and radiance such that the radiance parametric model is optimized without a dependency on the correct SDF. 

\paragraph{SDF regularization.}
A signed distance function $f(x)$ is differentiable almost everywhere, while its gradient $\grad f(x)$ satisfies the Eikonal equation $\|\grad f(x)\|_2 = 1$~.
This implies that an SDF can be trained with the Eikonal regularization. 
According to IGR~\cite{gropp2020implicit}, we enforce an Eikonal loss as a regularizer to make an implicit field act as a signed distance field. However, we discovered that training with a fixed Eikonal regularization~\cite{wang2021neus, wang2022hf, zhao2022human} can lead to a sub-optimal convergence where the SDF has converged based on the Eikonal regularization but the rendering can still be further optimized. 
Such a problem harms the improvement of the RGB rendering because the weights are projected from a wrong SDF value, which makes the extracted mesh miss the detailed structures.

\paragraph{Adaptive regularization for rendering.}
To solve the problem of the reconstruction failures on the thin structures, we found that the capability of rendering such structures from the radiance field should be ensured even when the SDF does not contain the accorded structures so that the gradient could be back-propagated from the radiance field to the SDF later on.
Based on the overall loss function $\mathcal{L}_\text{total} = \mathcal{L}_\text{rgb} + \mathcal{L}_\text{sdf}$~,
we propose to adaptively weight the Eikonal regularization to optimize the SDF as
\begin{align}
    \mathcal{L}_\text{sdf} = \frac{\lambda_\text{E}}{m n} \sum_{i=1}^{m} \lambda_\text{r}(r_i) \sum_{j=1}^{n} \left(\| \text{n}_{ij} \|_2 - 1 \right)^2 ~,
    \label{equ:loss_eikonal}
\end{align}
where a ray-wise weight $\lambda_\text{r}(r_i)$ for the $i$-th ray $r_i$ is set to be 
\begin{align}
    \lambda_\text{r}(r_i) = \frac{\alpha}{d_\text{r}(r_i) + \alpha} ~,
    \label{equ:lambda_r}
\end{align}
where $d_\text{r}(r_i)$ is the radiance distance $\| \mathbf{\hat{C}}(r_i) - \mathbf{C}(r_i) \|_2$ regarding ray $r_i$, and $\alpha$ is a positive hyperparameter that is set to be smaller than 1, \eg $1 \cdot 10^{-6}$. Consequently, the Eikonal regularization will be relaxed when the rendering quality determined by the metric of $d_\text{r}(r_i)$ is unsatisfying.
In our implementation, the approximated normal $\text{n} = \grad f(r(\cdot))$ is the derivative of $f(r(\cdot))$, and $\lambda_\text{E}$ is typically set to be 0.1 as mentioned by IGR~\cite{gropp2020implicit} and NeuS~\cite{wang2021neus}. 

To avoid extreme values with respect to different scenes, $d_\text{r}(r)$ is bounded to a range $[c_\text{min}, c_\text{max}]$. 
Notice that we end up with the typical Eikonal regularization proposed in IGR~\cite{gropp2020implicit} when we achieve satisfactory renderings measured by the small $d_\text{r}(r)$. Thus, in contrast to \cite{gropp2020implicit}, our approach is more adaptive to the changes in both the rendering and SDF optimizations; thus, relaxing the restrictions from the predefined parameters.

The proposed adaptive Eikonal regularization solves the problem in reconstructing detailed structures through the SDF like the wheel spokes of the bicycle, 
for which a strong Eikonal constraint will make them disappear as shown in \figref{fig:ablation_weight}~(a).
In contrast, our method was able to capture such structures in \figref{fig:ablation_weight}~(d).
Overall, \figref{fig:ablation_weight} illustrates that optimizing with the adaptive Eikonal loss has a better reconstruction of the circular shape of the wheels, the separation between the bench and the grass, and the holes between the blocks of wood on the bench.

\paragraph{Adaptive regularization to reduce geometric bias.}
Considering that a simple change in \eqref{equ:loss_eikonal} triggered a significant impact on the results. We propose to apply similar relaxing terms for the Eikonal regularization on each sampled point of the selected rays during training. Ideally, the space behind the observed geometric surface contributes far fewer points compared to the empty space in the neural rendering. Although this is a good phenomenon for efficient rendering, it also leads to a problem where an integral rendering point which could serve as an estimated depth point for each ray is biased by the ray's zero-crossing. As mentioned by D-NeuS~\cite{chen2023recovering}, the ideal SDF distribution is not guaranteed by the parametric geometric model. Although the parameters of the geometric model are explicitly initialized to produce a spherical SDF, the radiance-based supervision imposes no explicit regularization on the underlying SDF field. The inconsistency between the radiance field and SDF leads to the difficulty of optimizing the inner space in SDF, especially within tiny structures with a small negative space.

\begin{figure}[!t]
\centering
\includegraphics[width=\linewidth]
{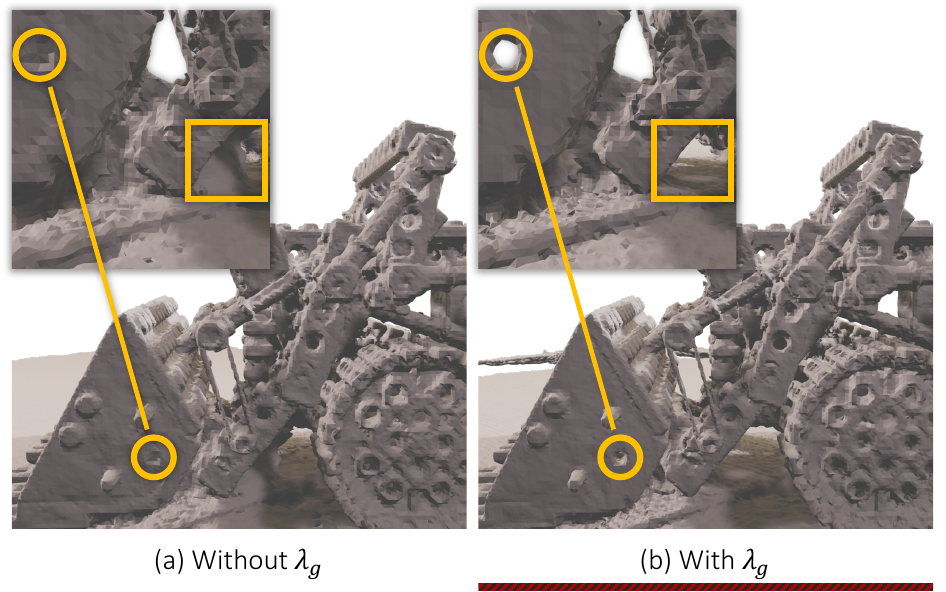}
\caption{Comparison of Neuralangelo trained with and without the proposed geometric bias factor $\lambda_\text{g}$ on top of the model trained with $\lambda_\text{r}$.}
\label{fig:lambda_g}
\end{figure}

Considering D-NeuS defines a weighted rendering point $r(t_\text{r})$ which is $ o~+~t_\text{r} \cdot v$ via discretizing and integrating the volume between $t_\text{n}$ and $t_\text{f}$ as $t_\text{r}$. 
Referring to the radiance $\mathbf{\hat{C}}(r)$ in \eqref{equ:radiance} of ray $r$ and the transmittance representation $T_\text{s}(t)$, our rendering point $t_\text{r}$ is weighted summed through $\omega(t)$ written as 
\begin{align}
    t_\text{r} = \frac{\sum_{j=1}^N \omega(t_j) \cdot t_j}{\sum_{j=1}^N \omega(t_j)} ~,
\end{align}
where the weight is represented as
\begin{align}
\omega(t_j) = \frac{\sigma(r(t_j))}{1+\exp\left({s \cdot f(r(t_j))}\right)} ~.
\end{align}
Given the calculated rendering point $t_\text{r}$, we introduce a geometric bias factor $\lambda_\text{g}$ which is
\begin{align}
    \lambda_\text{g} = 1 - \frac{t_\text{r} - t_\text{s}}{t_\text{f} - t_\text{n}} ~,
\end{align}
where $r(t_\text{s})$ is the approximated zero-crossing by assessing the SDF values along ray $r(\cdot)$. So, in this case, the Eikonal regularization will only be fully enforced when there is no geometric bias between the zero-crossing surface of the SDF and the rendering point in the radiance field. 
Weighted by $\lambda_\text{g}(r)$, the SDF loss also back-propagates gradients to the factor $s$ used in $T_\text{s}(t)$ which adjusts the projection from the SDF to the radiance field. 
Eventually, our final geometric loss is defined as 
\begin{align}
    \mathcal{L}_\text{sdf} = \frac{\lambda_\text{E}}{m n} \sum_{i=1}^{m} \lambda_\text{g}(r_i) \lambda_\text{r}(r_i) \sum_{j=1}^{n} \left(\| \text{n}_{ij} \|_2 - 1 \right)^2 ~.
\end{align}
As shown in ~\figref{fig:lambda_g}, our model trained with the additional geometric bias factor $\lambda_\text{g}(r)$ successfully reveals the tiny hole in the lifter of the Lego bulldozer in Mip-NeRF 360~\cite{barron2022mip} dataset.

\begin{figure*}[!ht]
\centering
\includegraphics[width=\linewidth]{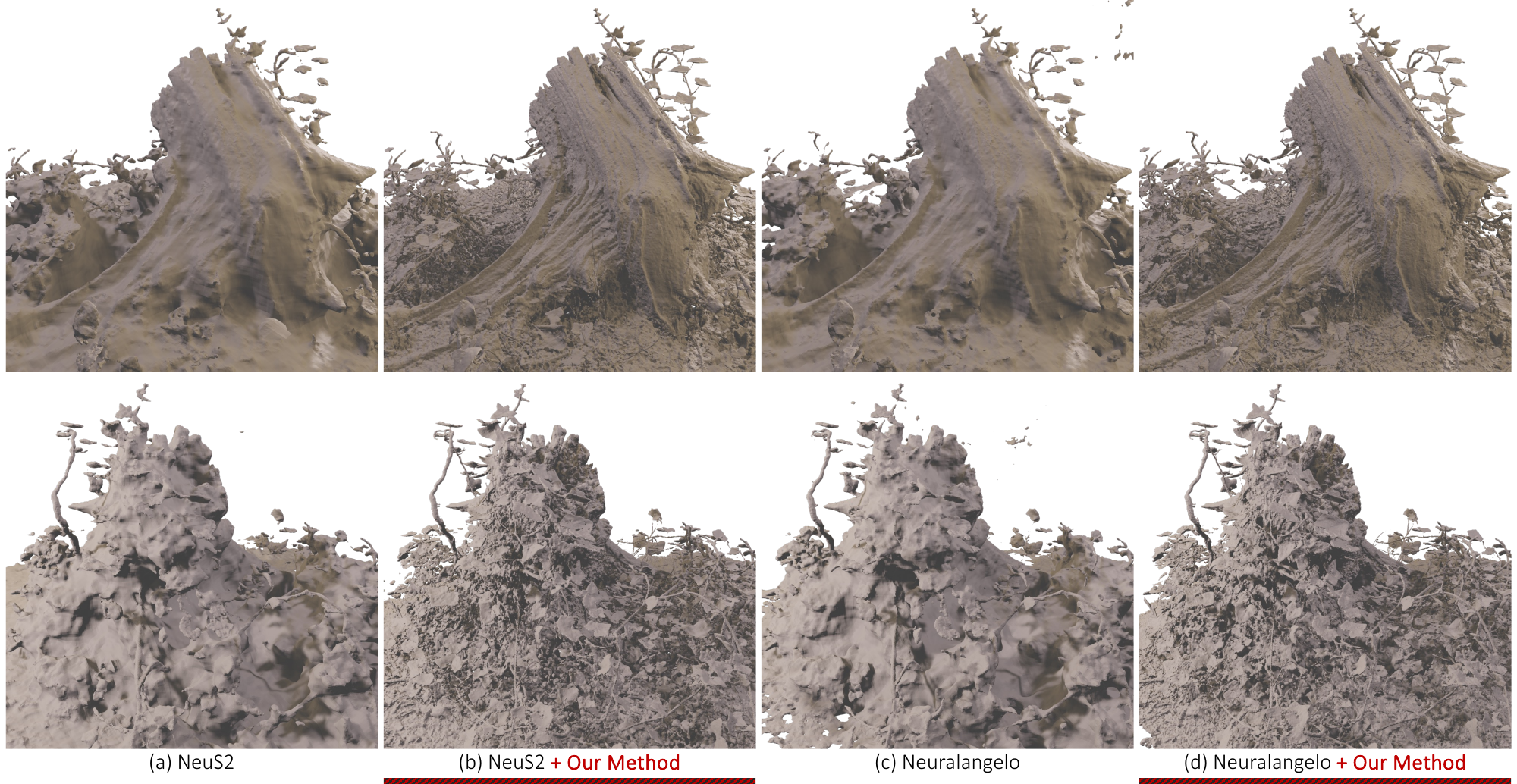}
\caption{Geometric reconstruction comparison evaluated on the Mip-NeRF 360 dataset~\cite{barron2022mip}.}
\label{fig:qualitative_mip_nerf}
\end{figure*}

\begin{figure}[!ht]
\centering
\includegraphics[width=\linewidth]{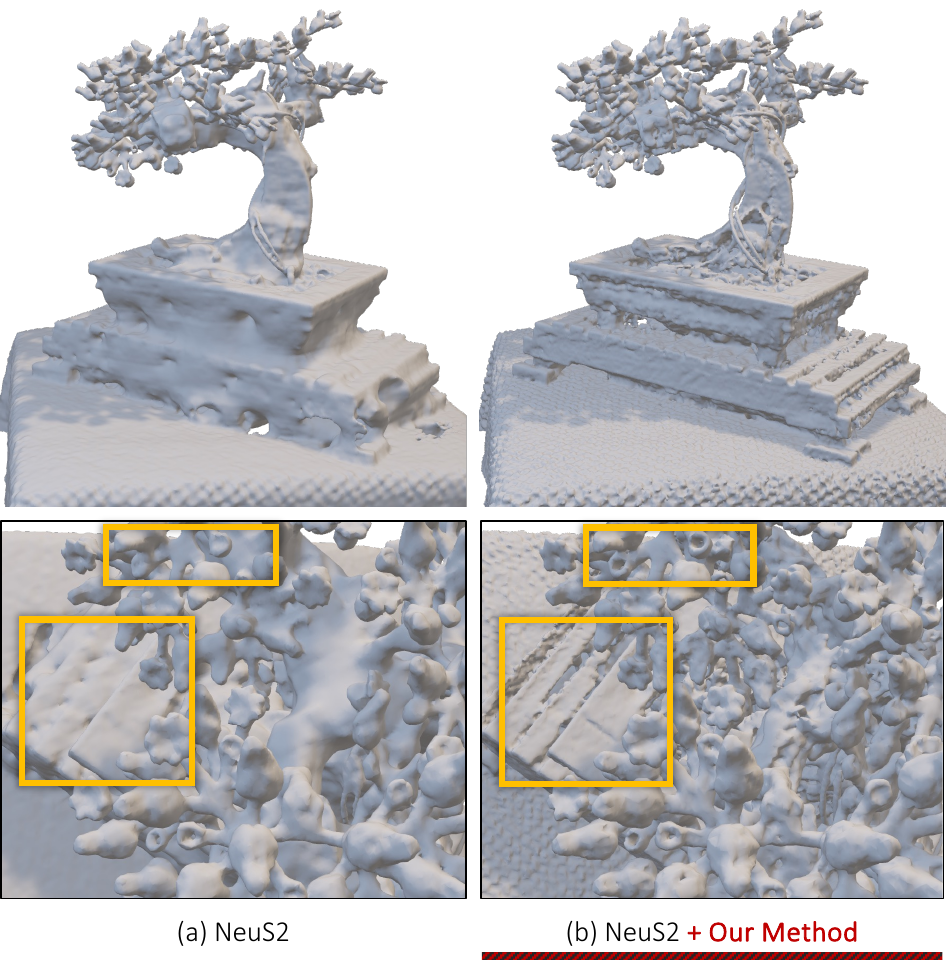}
\caption{Comparison of our mesh to Neus~2~\cite{wang2022neus2}, focusing on some important details on the bonsai dataset that our method was able to reconstruct while NeuS~2 missed.}
\label{fig:mesh_improved}
\end{figure}

\subsection{Training}
Our model is trained to simultaneously run the geometric optimization in 3D space and appearance optimization in 2D space with $\mathcal{L}_\text{total}$.  
The Adam optimizer~\cite{kingma2014adam} is adopted with an exponentially decaying learning rate schedule ranging from $1\cdot10^{-2}$ to $1\cdot10^{-4}$. 

Trained with the proposed weighting factors $\lambda_\text{r}$ and $\lambda_\text{g}$ where $\alpha$ is set to be $1 \cdot 10^{-6}$ in \eqref{equ:lambda_r}, a generalizable model \textit{RaNeuS} can be built on top of NeuS~2~\cite{wang2022neus2} with a customized background model. 
Given $x$, a hash encoding is obtained at each level $h_i(x) \in \mathbb{R}^d$, where $d$ is the dimension of a feature vector and $i = 1, \dots, L$. The hash codes $h_i(x)$ are presented by interpolating the feature vectors assigned at the surrounding voxel grids at level $i$. Consequently, the hash encodings at all $L$ levels are then concatenated into the multi-resolution hash encoding $h(x) = \{h_i(x)\}^L_{i=1} \in \mathbb{R}^{L \times d}$.
Our base resolution in the hash table is set to be $32\times32\times32$ and eventually reaches a resolution of $2048\times2048\times2048$ on the 16th level. Progressive training is applied to stabilize the low-level hash table during training, where the first 4 levels are optimized together in the first 2,000 steps and a new higher level is included for optimization every other 2,000 steps.  

To deploy \textit{RaNeuS} in unbounded scenes, we represent the foreground and background space in two separate models. The foreground space is modeled by NeuS~2 in a cubic bounding box which covers the scene at close proximity; while, the background is modeled by a vanilla NeRF for rendering alone which is encoded by MLP without an SDF intermediate, which is learned in a contracted spherical space with a radius of twice the diagonal length of the cuboid. 


\begin{table*}[!ht]
\centering
{
\begin{tabular}{lccccccccc|c}
\toprule
 Methods 
 & bicycle & flowers & garden & stump & treehill & room & counter & kitchen & bonsai & \textit{Avg.} \\
\midrule 
NeRF~\cite{mildenhall2021nerf} & 21.76 & 19.40 & 23.11 & 21.73 & 21.28 & 28.56 & 25.67 & 26.31 & 26.81 & 23.85 \\
Mip-NeRF~\cite{barron2021mip} & 21.69 & 19.31 & 23.16 & 23.10 & 21.21 & 28.73 & 25.59 & 26.47 & 27.13 & 24.04 \\
NeRF++~\cite{zhang2020nerf++} & 22.64 & 20.31 & 24.32 & 24.34 & 22.20 & 28.87 & 26.38 & 27.80 & 29.15 & 25.11 \\
Deep Blending~\cite{hedman2018deep} & 21.09 & 18.13 & 23.61 & 24.08 & 20.80 & 27.20 & 26.28 & 25.02 & 27.08 & 23.70 \\
Point-Based~\cite{kopanas2021point} & 21.64 & 19.28 & 22.50 & 23.90 & 20.98 & 26.99 & 25.23 & 24.47 & 28.42 & 23.71 \\
Mip-NeRF 360~\cite{barron2022mip} & 24.37 & 21.73 & 26.98 & {26.40} & 22.87 & {31.63} & 29.55 & \textbf{32.23} & 33.46 & 27.69 \\
NeRF2Mesh~\cite{tang2023delicate} & 22.44 & -- & 23.13 & 23.53 & -- & 28.19 & 24.11 & 23.75 & 24.34 & -- \\
\midrule 
HF-NeuS~\cite{wang2022hf} & 23.99 & 21.16 & 26.19 & 25.26 & 21.50 & 30.07 & 29.14 & 29.70 & 34.08 & 26.78 \\
+ \textit{Proposed Method} & \textbf{25.40} & \textbf{22.92} & \textbf{27.65} & \textbf{26.63} & \textbf{23.20} & \textbf{31.80} & \textbf{30.53} & 31.29 & \textbf{35.75} & \textbf{28.35} \\

\bottomrule
\end{tabular}
}
\caption{Mean PSNR on different scenes in Mip-NeRF 360 dataset~\cite{barron2022mip}. HF-NeuS~\cite{wang2022hf} is also parameterized with a background NeRF~\cite{mildenhall2021nerf} model for rendering alone to cover the whole unbounded scene. 
 \label{tab:psnr_mipnerf360}
}

\end{table*}

\begin{table*}[!t]
\centering
{
\begin{tabular}{lccccccccc|c}
\toprule
 Methods 
 & bicycle & flowers & garden & stump & treehill & room & counter & kitchen & bonsai & \textit{Avg.} \\
\midrule 
NeRF~\cite{mildenhall2021nerf} & 0.455 & 0.376 & 0.546 & 0.453 & 0.459 & 0.843 & 0.775 & 0.749 & 0.792 & 0.605 \\
Mip-NeRF~\cite{barron2021mip} & 0.454 & 0.373 & 0.543 & 0.517 & 0.466 & 0.851 & 0.779 & 0.745 & 0.818 & 0.616 \\
NeRF++~\cite{zhang2020nerf++} & 0.526 & 0.453 & 0.635 & 0.594 & 0.530 & 0.852 & 0.802 & 0.816 & 0.876 & 0.676 \\
Deep Blending~\cite{hedman2018deep} & 0.466 & 0.320 & 0.675 & 0.634 & 0.523 & 0.868 & 0.856 & 0.768 & 0.883 & 0.666 \\
Point-Based~\cite{kopanas2021point} & 0.608 & 0.487 & 0.735 & 0.651 & 0.579 & 0.887 & 0.868 & 0.876 & 0.919 & 0.734 \\
Mip-NeRF 360~\cite{barron2022mip} & 0.685 & 0.583 & 0.813 & 0.744 & 0.632 & {0.913} & {0.894} & 0.920 & 0.941 & 0.792 \\
\midrule 
HF-NeuS~\cite{wang2022hf} & 0.512 & 0.547 & 0.747 & 0.651 & 0.585 & 0.739 & 0.855 & 0.726 & 0.824 & 0.687 \\
+ \textit{Proposed Method} & \textbf{0.721} & \textbf{0.641} & \textbf{0.870} & \textbf{0.797} & \textbf{0.688} & \textbf{0.942} & \textbf{0.926} & \textbf{0.955} & \textbf{0.951} & \textbf{0.832} \\

\bottomrule
\end{tabular}
}
\caption{Structural similarity index measure (SSIM)~\cite{wang2004image, brunet2011mathematical} on different scenes in Mip-NeRF 360 dataset~\cite{barron2022mip}. 
 \label{tab:ssim_mipnerf360}
}
\end{table*}

\section{Experiments}

Since the proposed approach can be adapted to the existing methods, we demonstrate the advantage of our contributions by integrating them with NeuS~2~\cite{wang2022neus2}, HF-NeuS~\cite{wang2022hf} and the more recent Neuralangelo~\cite{li2023neuralangelo}. Both works can benefit from the proposed ray-adaptive factors $\lambda_\text{r}$ and $\lambda_\text{g}$.
We conduct our evaluation on three datasets: 
Mip-NeRF~\cite{barron2021mip},
NeRF-synthetic~\cite{mildenhall2021nerf} and
DTU~\cite{jensen2014large}.
To evaluate the quality of the reconstruction, Chamfer distance is used for 3D geometric evaluation, and PSNR is used for rendering validation.

\subsection{Mip-NeRF 360}

The Mip-NeRF 360 dataset~\cite{barron2022mip} provides nine unbounded scenes, which include five outdoor and four indoor scenes, each containing a complex central object or area and a detailed background. They relied on COLMAP to estimate the camera poses while the camera intrinsics are shared among all images in a scene. The color harmonization issue is limited here, where the captured outdoor scene is collected when the sky is overcast, ensuring that the camera operator casts soft shadows that minimally affect the illumination in the scene. For the indoor scenes, they relied on large diffuse light sources. 
Each scene has between 100 to 330 images with a resolution of 1.0–1.6 megapixels. 


We compare the rendering performance with PSNR in \tabref{tab:psnr_mipnerf360} and SSIM in \tabref{tab:ssim_mipnerf360}. Our model achieves the best overall results of 28.35 for PSNR and 0.832 for SSIM, while Mip-NeRF 360~\cite{barron2022mip} performs better in a few categories. 

In terms of geometry, we highlight in \figref{fig:mesh_improved} the amount of detail our results can handle compared to NeuS~2~\cite{wang2022neus2}.
Particularly, the over-smoothed NeuS~2 reconstruction of the bonsai is because of the shadows and changes in exposures. The improvement demonstrates that our contributions help overcome this problem. We also include more results against the Neus~2~\cite{wang2022neus2} and Nerualangelo~\cite{li2023neuralangelo} in \figref{fig:qualitative_mip_nerf} where we illustrate the advantage of the proposed method to reconstruct the finer details.

\begin{table}[!t]
\centering
\resizebox{0.99\linewidth}{!}
{
\begin{tabular}{lcccccc|c}
\toprule
 Methods 
 & Chair & Ficus & Lego & Mat. & Mic & Ship & \textit{Avg.} \\
\midrule 
NeRF~\cite{mildenhall2021nerf} & 33.00 & 30.15 & 32.54 & 29.62 & 32.91 & 28.34 & 31.09 \\
Mip-NeRF~\cite{barron2021mip} & \textbf{37.14} & {33.18} & \textbf{35.74} & \textbf{32.56} & \textbf{38.04} & \textbf{33.08} & \textbf{34.96} \\
\midrule 
VolSDF~\cite{yariv2021volume} & 25.91 & 24.41 & 26.99 & 28.83 & 29.46 & 25.65 & 26.86 \\
NeuS~\cite{wang2021neus} & 27.95 & 25.79 & 29.85 & 29.36 & 29.89 & 25.46 & 28.05 \\
Instant-NSR~\cite{zhao2022human} & 34.04 & 32.47 & 33.78 & 27.67 & 33.43 & 29.50 & 31.81 \\
\midrule 
HF-NeuS~\cite{wang2022hf} & 28.69 & 26.46 & 30.72 & 29.87 & 30.35 & 25.87 & 28.66 \\
+ \textit{Proposed Method} & 35.26 & \textbf{34.02} & 34.51 & 28.99 & 35.51 & 33.02 & 33.55 \\

\bottomrule
\end{tabular}
}
\caption{Mean PSNR on different scenes in NeRF-synthetic dataset~\cite{mildenhall2021nerf}. Note that the upper parts in this table are methods focusing on rendering alone, while the lower parts in this table contain geometric regularization during training.
 \label{tab:psnr_synthetic}
}
\end{table}

\subsection{NeRF-synthetic}
NeRF-synthetic dataset~\cite{mildenhall2021nerf} contains objects with fine-grained detailed and sharp features, such as the Lego bulldozer and a model ship. This dataset was first validated by NeRF~\cite{mildenhall2021nerf} for volume rendering. Since it consists entirely of synthetic objects in front of a white background, the rendering task is easier compared to unbounded scenes such as Mip-NeRF 360 dataset~\cite{barron2022mip}.

We first compare against NeRF~\cite{mildenhall2021nerf} and Mip-NeRF~\cite{barron2021mip} in \tabref{tab:psnr_synthetic} which are methods that focus on novel view synthesis and do not necessarily extract any mesh.
Although the proposed method is not the state-of-the-art compared to Mip-NeRF, the gap is noticeably small. 
Among methods that jointly optimize for geometry and appearance,  
we demonstrate improved rendering accuracy compared to other methods, achieving the best results. 
To highlight the improved geometries, we present the qualitative results in \figref{fig:teaser}, the ropes connecting the poles and the decker are more complete compared to 
HF-NeuS~\cite{wang2022hf} and Neuralangelo~\cite{li2023neuralangelo}.

Considering that NeRF-synthetic is a comparably small dataset, convergence is easily achieved. For this evaluation, all models are coded with hash coding to converge within 20 minutes.

\begin{table*}[!t]
\centering
\resizebox{\linewidth}{!}
{
\begin{tabular}{lccccccccccccccc|c}
\toprule
 Methods 
 & 24 & 37 & 40 & 55 & 63 & 65 & 69 & 83 & 97 & 105 & 106 & 110 & 114 & 118 & 122 & \textit{Avg.} \\
\midrule 
NeRF~\cite{mildenhall2021nerf} & 26.24 & 25.74 & 26.79 & 27.57 & 31.96 & 31.50 & 29.58 & 32.78 & 28.35 & 32.08 & 33.49 & 31.54 & 31.00 & 35.59 & 35.51 & 30.65 \\
VolSDF~\cite{yariv2021volume} & 26.28 & 25.61 & 26.55 & 26.76 & 31.57 & 31.50 & 29.38 & 33.23 & 28.03 & 32.13 & 33.16 & 31.49 & 30.33 & 34.90 & 34.75 & 30.38 \\
NeuS~\cite{wang2021neus} &  28.20 & 27.10 & 28.13 & 28.80 & 32.05 & 33.75 & 30.96 & 34.47 & 29.57 & 32.98 & 35.07 & 32.74 & 31.69 & 36.97 & 37.07 & 31.97 \\
D-NeuS~\cite{chen2023recovering} & 28.98 & 27.58 & 28.40 & 28.87 & 33.71 & 33.94 & 30.94 & 34.08 & 30.75 & 33.73 & 34.84 & 32.41 & 31.42 & 36.76 & 37.17 & 32.22 \\
\midrule 
HF-NeuS~\cite{wang2022hf} & 29.15 & 27.33 & 28.37 & 28.88 & 32.89 & 33.84 & 31.17 & 34.83 & 30.06 & 33.37 & 35.44 & 33.09 & 32.12 & 37.13 & 37.32 & 32.33 \\
+ \textit{Proposed Method} & 31.14 & 28.19 & 29.13 & 30.09 & 34.77 & 35.62 & \textbf{33.84} & 36.70 & \textbf{33.19} & 36.44 & 37.22 & \textbf{35.91} & {34.55} & 38.83 & 39.35 & 34.33 \\
- without $\lambda_\text{g}$ & 31.09 & 28.21 & 28.97 & {29.91} & {34.25} & {35.15} & 33.26 & {36.74} & 32.30 & 36.02 & 36.96 & 35.22 & \textbf{34.70} & {38.29} & 38.93 & 34.00 \\
- without $\lambda_\text{r}$ & 30.01 & 28.03 & 28.41 & 28.97 & 33.12 & 34.15 & 32.28 & 35.21 & 30.94 & 34.55 & 35.91 & 34.22 & 33.49 & 37.82 & 38.29 & 33.03 \\
\midrule 
Neuralangelo~\cite{li2023neuralangelo} & 30.64 & 27.78 & 32.70 & 34.18 & 35.15 & 35.89 & 31.47 & 36.82 & 30.13 & 35.92 & 36.61 & 32.60 & 31.20 & 38.41 & 38.05 & 33.84 \\
+ \textit{Proposed Method} & \textbf{32.31} & \textbf{29.71} & \textbf{35.11} & 35.96 & \textbf{37.57} & \textbf{37.71} & 33.37 & \textbf{38.35} & 32.14 & \textbf{38.10} & \textbf{38.90} & 33.93 & 33.41 & 40.24 & \textbf{39.60} & \textbf{35.76} \\
- without $\lambda_\text{g}$ & 32.15 & 29.25 & 34.72 & \textbf{36.21} & 37.41 & 37.21 & 33.6 & 38.24 & 31.59 & 37.51 & 38.44 & 34.13 & 32.65 & \textbf{40.60} & 39.37 & 35.54 \\
- without $\lambda_\text{r}$ & 31.65 & 29.02 & 34.14 & 35.24 & 36.59 & 37.08 & 32.4 & 37.73 & 31.14 & 37.10 & 37.89 & 33.27 & 32.67 & 39.44 & 38.74 & 34.94 \\

\bottomrule
\end{tabular}
}
\caption{Mean PSNR on different objects in DTU dataset~\cite{jensen2014large}.
 \label{tab:psnr_dtu}
}
\end{table*}

\begin{table*}[!t]
\centering
\resizebox{\linewidth}{!}
{
\begin{tabular}{lccccccccccccccc|c}
\toprule
 Methods 
 & 24 & 37 & 40 & 55 & 63 & 65 & 69 & 83 & 97 & 105 & 106 & 110 & 114 & 118 & 122 & \textit{Avg.} \\
\midrule 
COLMAP~\cite{schoenberger2016mvs} & 0.81 & 2.05 & 0.73 & 1.22 & 1.79 & 1.58 & 1.02 & 3.05 & 1.40 & 2.05 & 1.00 & 1.32 & 0.49 & 0.78 & 1.17 & 1.36 \\
Instant-NGP~\cite{muller2022instant} & 1.68 & 1.93 & 1.57 & 1.16 & 2.00 & 1.56 &  1.81 & 2.33 & 2.16 & 1.88 & 1.76 & 2.32 & 1.86 & 1.80 & 1.72 & 1.84 \\
IDR~\cite{yariv2020multiview} & 1.63 & 1.87 & 0.63 & 0.48 & 1.04 & 0.79 & 0.77 & 1.33 & 1.16 & 0.76 & 0.67 & 0.90 & 0.42 & 0.51 & 0.53 & 0.90 \\
MVSDF~\cite{zhang2021learning} & 0.83 & 1.76 & 0.88 & 0.44 & 1.11 & 0.90 & 0.75 & 1.26 & 1.02 & 1.35 & 0.87 & 0.84 & 0.34 & 0.47 & 0.46 & 0.88 \\
RegSDF~\cite{zhang2022critical} & 0.60 & 1.41 & 0.64 & 0.43 & 1.34 & 0.62 & 0.60 & \textbf{0.90} & 0.92 & 1.02 & 0.60 & \textbf{0.60} & 0.30 & 0.41 & 0.39 & 0.72 \\
NeRF~\cite{mildenhall2021nerf} & 1.90 & 1.60 & 1.85 & 0.58 & 2.28 & 1.27 & 1.47 & 1.67 & 2.05 & 1.07 & 0.88 & 2.53 & 1.06 & 1.15 & 0.96 & 1.49 \\
VolSDF~\cite{yariv2021volume} & 1.14 & 1.26 & 0.81 & 0.49 & 1.25 & 0.70 & 0.72 & 1.29 & 1.18 & 0.70 & 0.66 & 1.08 & 0.42 & 0.61 & 0.55 & 0.86 \\
NeuS~\cite{wang2021neus} & 1.00 & 1.37 & 0.93 & 0.43 & 1.10 & 0.65 & 0.57 & 1.48 & 1.09 & 0.83 & 0.52 & 1.20 & 0.35 & 0.49 & 0.54 & 0.84 \\ 
NeuralWarp~\cite{darmon2022improving} & 0.49 & 0.71 & 0.38 & 0.38 & 0.79 & 0.81 & 0.82 & 1.20 & 1.06 & 0.68 & 0.66 & 0.74 & 0.41 & 0.63 & 0.51 & 0.68 \\
D-NeuS~\cite{chen2023recovering} & 0.44 & 0.79 & 0.35 & 0.39 & 0.88 & 0.58 & 0.55 & 1.35 & 0.91 & 0.76 & \textbf{0.40} & 0.72 & {0.31} & 0.39 & 0.39 & 0.61 \\
\midrule 
HF-NeuS~\cite{wang2022hf} & 0.76 & 1.32 & 0.70 & 0.39 & 1.06 & 0.63 & 0.63 & 1.15 & 1.12 & 0.80 & 0.52 & 1.22 & 0.33 & 0.49 & 0.50 & 0.77 \\
+ \textit{Proposed Method} & 0.50 & 0.81 & \textbf{0.26} & \textbf{0.26} & 0.75 & 0.53 & 0.54 & 0.93 & \textbf{0.79} & 0.64 & 0.41 & {0.62} & 0.30 & 0.35 & \textbf{0.24} & 0.53 \\
- without $\lambda_\text{g}$ & 0.51 & 0.82 & 0.36 & 0.34 & 0.81 & 0.55 & 0.56 & {0.98} & 0.87 & {0.66} & 0.43 & 0.71 & 0.31 & 0.40 & {0.34} & 0.58 \\
- without $\lambda_\text{r}$ & 0.68 & 1.15 & 0.58 & 0.37 & 0.92 & 0.59 & 0.64 & 1.06 & 0.99 & 0.75 & 0.46 & 0.93 & 0.33 & 0.45 & 0.41 & 0.69 \\
\midrule 
Neuralangelo~\cite{li2023neuralangelo} & 0.37 & 0.72 & 0.35 & 0.35 & 0.87 & 0.54 & 0.53 & 1.29 & 0.97 & 0.73 & 0.47 & 0.74 & 0.32 & 0.41 & 0.43 & 0.61 \\
+ \textit{Proposed Method} & 0.33 & 0.65 & 0.33 & 0.32 & 0.90 & \textbf{0.44} & \textbf{0.46} & 1.32 & 0.90 & 0.60 & \textbf{0.40} & {0.67} & \textbf{0.27} & 0.29 & 0.36 & 0.55 \\
- without $\lambda_\text{g}$ & 0.35 & 0.67 & 0.33 & 0.32 & 0.88 & 0.49 & 0.48 & 1.31 & 0.90 & 0.68 & 0.41 & 0.68 & 0.34 & {0.37} & 0.39 & 0.57 \\
- without $\lambda_\text{r}$ & 0.37 & 0.72 & 0.33 & 0.36 & 0.91 & 0.51 & 0.52 & 1.35 & 0.93 & 0.69 & 0.45 & 0.70 & 0.36 & 0.38 & 0.42 & 0.60 \\
\midrule
NeuS~2~\cite{wang2022neus2} & 0.56 & 0.76 & 0.49 & 0.37 & 0.92 & 0.71 &  0.76 & 1.22 & 1.08 & 0.63 & 0.59 & 0.89 & 0.40 & 0.48 & 0.55 & 0.70 \\
+ \textit{Proposed Method} & \textbf{0.31} & \textbf{0.59} & 0.29 & 0.28 & \textbf{0.74} & 0.45 &  0.51 & 1.01 & 0.82 & \textbf{0.59} & 0.41 & 0.73 & 0.39 & \textbf{0.28} & 0.29 & \textbf{0.51} \\

\bottomrule
\end{tabular}
}
\caption{Fidelity which is reported using Chamfer distance~(mm) on different objects in the DTU dataset~\cite{jensen2014large}.
 \label{tab:fid_dtu}
}
\end{table*}

\subsection{DTU}

\begin{figure}[!t]
\centering
\includegraphics[width=\linewidth]{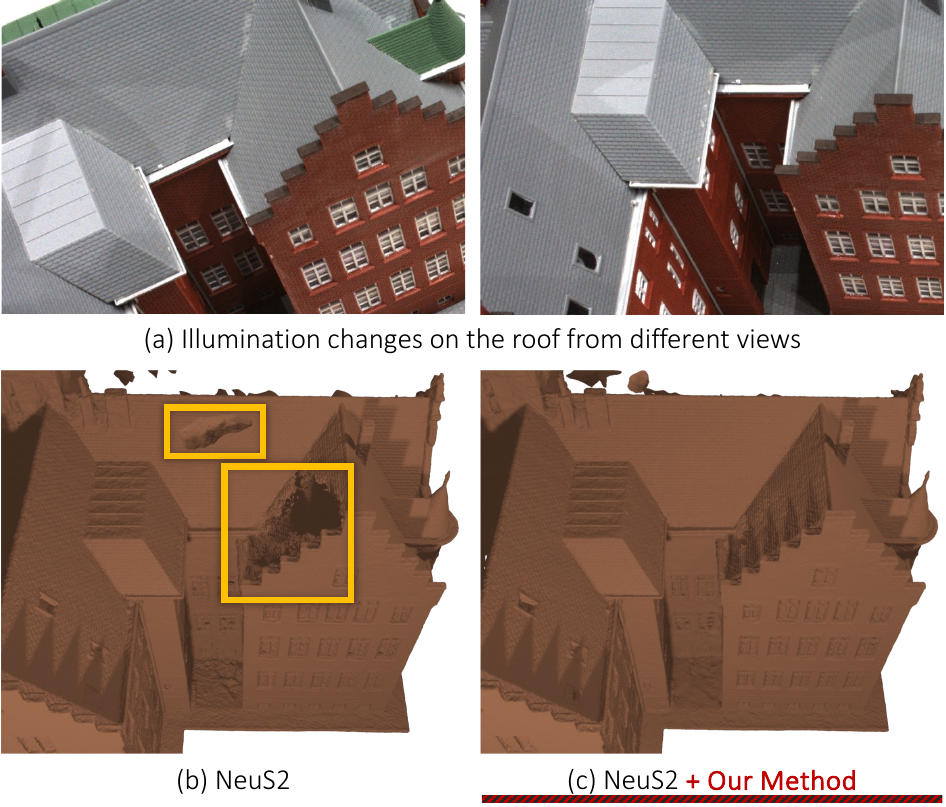}
\caption{Our adaptive training makes the learned geometry 
robust against the shadows triggered by the movement of the light source.
}
\label{fig:shadow}
\end{figure}

DTU~\cite{jensen2014large} is a multi-view stereo dataset. We selected the same 15 models for comparison as the previous methods. 
Each scene consists of 49 or 64 views with 1600 $\times$ 1200 resolution. As previously mentioned, we use Chamfer distance to measure the accuracy and completeness. 

Similar to the previous works, we clean the extracted meshes with the object masks dilated by 50 pixels. By doing so, we have achieved better results for both rendering PSNR and mesh reconstruction without using any 3D supervision, such as sparse or dense depth. 
On average, we reached the state-of-the-art in \tabref{tab:psnr_dtu} and \tabref{tab:fid_dtu}. Note that our approach even outperforms RegSDF~\cite{zhang2022critical} which was trained with 3D supervision.


\paragraph{Ablation study.}
In these tables, we included an ablation study to highlight the advantage of each contribution on HF-NeuS and Neuralangelo. \tabref{tab:psnr_dtu} and \tabref{tab:fid_dtu} show that both contributions increase PSNR and reduce the Chamfer distance. Our method shows more robustness against the illumination change among different views as shown in \figref{fig:shadow}.

\section{Conclusion}

We propose reconstructing a mesh of a static scene or an object trained with the novel adaptive Eikonal regularization. 
%
%
Compared to the related works, 
the evaluations on the synthetic and real datasets prove that our model performs the best on novel view synthesis among neural renderers, with implicit fields as intermediate representations. In terms of the 3D reconstruction, our evaluation on the DTU dataset shows that the proposed method has the best and second-best geometric reconstruction accuracy. The thin structures in training images are utilized precisely as visual cues to optimize the SDF in 3D space.

\bibliographystyle{ACM-Reference-Format}
\bibliography{main}


\end{document}